\documentclass[eat,twocolumn]{jmlr}

\usepackage{longtable}% for long tables

\usepackage{booktabs}
\usepackage{multirow}
\usepackage{siunitx}
\usepackage{lipsum}
\usepackage{caption}
\usepackage{xcolor}
\definecolor{forestgreen}{rgb}{0.13, 0.55, 0.13}

\newcommand{\mypara}[1]{\vspace{1mm}\noindent\textbf{#1}}

\makeatletter
\DeclareRobustCommand\onedot{\futurelet\@let@token\@onedot}
\def\@onedot{\ifx\@let@token.\else.\null\fi\xspace}

\def\eg{\emph{e.g}\onedot} 
 
\def\cf{\emph{cf}\onedot} 
 \def\vs{\emph{vs}\onedot}

\makeatother

\theorembodyfont{\upshape}
\theoremheaderfont{\scshape}
\theorempostheader{:}
\theoremsep{\newline}

\jmlrvolume{}
\firstpageno{1}
\jmlryear{2022}
\jmlrworkshop{Machine Learning for Health (ML4H) 2022}

\title[SSL pretraining for CXR]{Can we Adopt Self-supervised Pretraining for Chest X-Rays?}

\author{
\Name{Arsh Verma} \Email{arsh@wadhwaniai.org} \and \\
\Name{Makarand Tapaswi} \Email{makarand@wadhwaniai.org}\\
\addr Wadhwani Institute for Artificial Intelligence (Wadhwani AI), India}

\begin{document}

\maketitle

\begin{abstract}
% Chest radiographs (or Chest X-rays, CXRs) are one of the most common forms of medical imaging modalities that are used by radiologists across the world to infer a disease diagnosis.
Chest radiograph (or Chest X-Ray, CXR) is a popular medical imaging modality that is used by radiologists across the world to diagnose heart or lung conditions.
Over the last decade, Convolutional Neural Networks (CNN), have seen success in identifying pathologies in CXR images.
% and outperform models trained from scratch.
Typically, these CNNs are pretrained on the standard ImageNet classification task, but this assumes availability of large-scale annotated datasets.
In this work, we analyze the utility of pretraining on \emph{unlabeled} ImageNet or Chest X-Ray (CXR) datasets using various algorithms and in multiple settings.
% \MT{maybe revisit}
Some findings of our work include:
(i)~supervised training with labeled ImageNet learns strong representations that are hard to beat;
(ii)~self-supervised pretraining on ImageNet ($\sim$1M images) shows performance similar to self-supervised pretraining on a CXR dataset ($\sim$100K images); and
(iii)~the CNN trained on supervised ImageNet can be trained further with self-supervised CXR images leading to improvements, especially when the downstream dataset is on the order of a few thousand images.
% is comparable to 
% and show that completely self-supervised (SSL) pretraining leads to performance comparable with labeled ImageNet.
% This opens up possibilities as obtaining large unlabeled data collections in the medical domain is easier and cost-effective.
% that training with large unlabeled data collections, may lead to better performance than models pretrained on ImageNet in a supervised setting.
\end{abstract}

\section{Introduction}
\label{sec:intro}

%% why x-rays and ML
Modern medicine uses advanced medical imaging techniques to assist physicians in disease diagnosis.
In particular, chest radiography (or Chest X-Rays, CXR) is a popular modality to identify pathologies in the lungs or heart owing to its low cost and decent availability~\citep{ngoya2016defining, smith2012use}.
Unfortunately, the number of X-Rays performed every year are rapidly rising, while the number of skilled radiologists required to analyze them is not keeping pace \citep{rsna_radiologists_shortage}.

\begin{figure}[t]
\centering
\includegraphics[width=\linewidth]{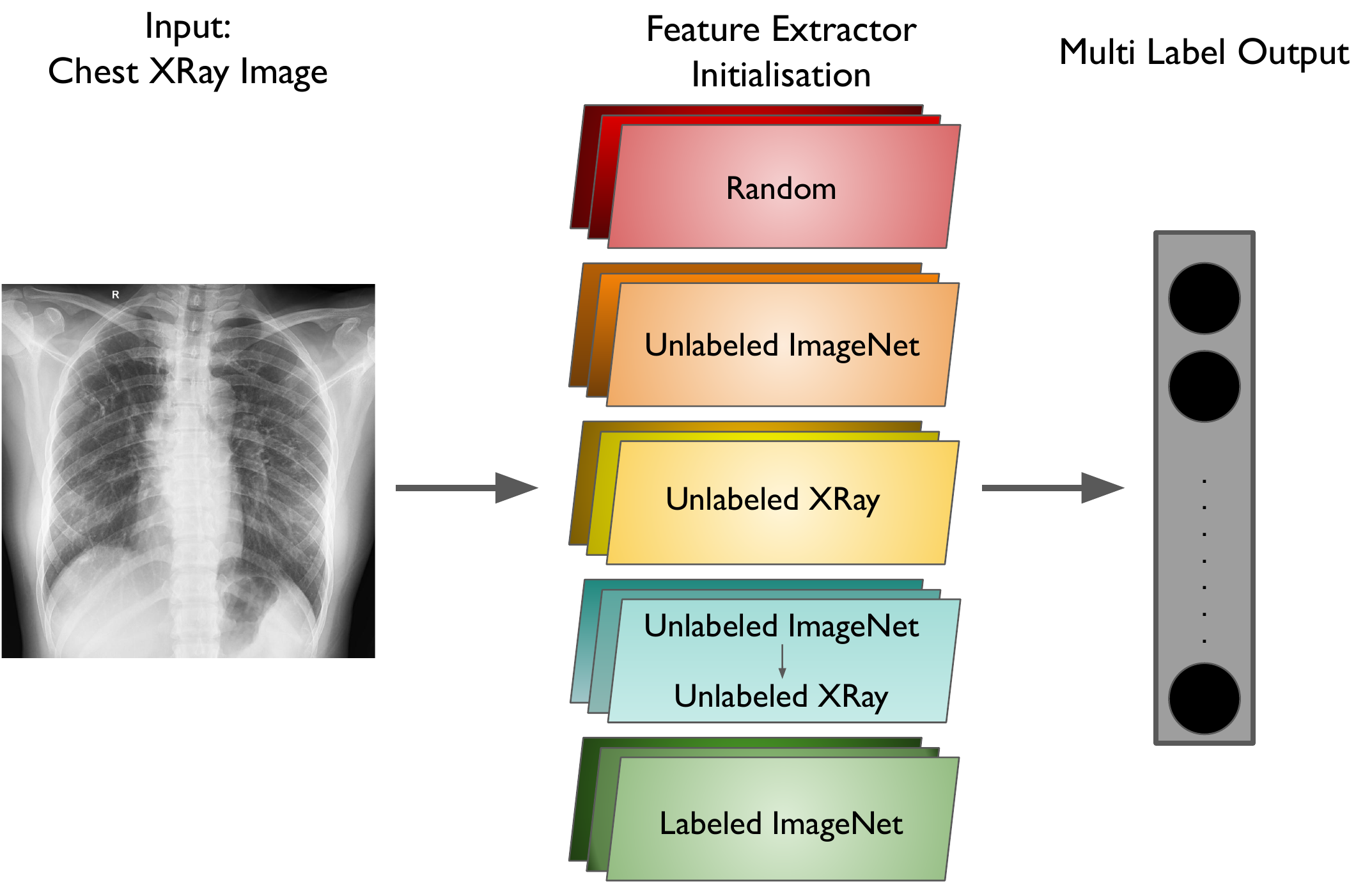}
\vspace{-4mm}
\caption{Overview of various training or initialization strategies adopted in our work for multi-label classification in Chest X-rays.}
\label{fig:overview}
\vspace{-4mm}
\end{figure}

%% challenges with classical supervised ML
This has spurred interest among Machine Learning researchers to develop models for automated detection of pathologies (\eg~\emph{consolidation}) in Chest X-Rays~\citep{rajpurkar2017chexnet, lakhani2017deep, tang2020automated}.
However, today's deep models are data hungry and while there is an abundance of X-Ray images, labeling them is a severe bottleneck.
Obtaining gold standard labels requires hiring several radiologists, who may still disagree with each other~\citep{albaum1996interobserver,johnson2010intraobserver}, leading to potentially erroneous labels~\citep{brady2017error}.
Such large datasets may also be annotated automatically with NLP parsers based on radiological reports~\citep{johnson2019mimic, irvin2019chexpert},
have varying prevalence for labels, and
may also have differing imaging properties.
Finally, relying on a set of annotations limits predictions to a closed set of labels, which are often different across datasets (\cf~Appendix~\ref{apd:first}).

%% ImageNet --> unlabeled data
A common practice to reduce the requirement of large labeled in-domain datasets is \emph{transfer~learning}
where models pretrained on one task are used for initializing models on the target task~\citep{raghu2019transfusion, matsoukas2022makes, sellergren2022simplified}.
With over a million images spanning thousand classes, ImageNet~\citep{deng2009imagenet} has emerged as the \emph{de facto} pretraining dataset for vision tasks and is surprisingly effective even for medical imaging problems.

As an alternative to labeling large-scale image datasets, self-supervised learning (SSL) has shown promise in learning features on a proxy task such as distinguishing a paired augmentation of the same image against other images~\citep{chen2020simple}.

\mypara{Findings.}
As illustrated in Fig.~\ref{fig:overview}, we wish to investigate the effect of using SSL pretraining (PT) on in-domain CXR \vs out-of-domain datasets.
We denote SSL~PT to mean pretraining with a self-supervised objective.
We conduct such experiments on two CXR datasets (see Table~\ref{tab:datasets}) and identify three key takeaways:
(i)~SSL~PT on the ImageNet or NIH dataset outperforms training from scratch by a large gap;
(ii)~SSL~PT on ImageNet ($\sim$1M images) achieves comparable performance to SSL PT on the NIH dataset ($\sim$100K images); and
(iii)~SSL~PT on ImageNet followed by SSL PT on the in-domain NIH improves downstream performance.

We also compare SSL against supervised learning, and find that SSL methods lag behind (sometimes only by a small margin) supervised ImageNet PT.
However, SSL~PT on in-domain CXR data provides small but consistent improvements to the supervised ImageNet representations.

Finally, we present two additional analyses:
(i)~the number of labels used for evaluation (5 \vs all) has a small influence on the gap between SSL PT \vs supervised ImageNet performance; and 

(ii)~zero-shot evaluation of models fine-tuned on one CXR dataset to another shows that SSL~PT learns robust models.

\mypara{Related SSL works in CXR.}
%% Recent work on SSL / contrastive learning
Recently, there have been significant efforts to explore related directions of
supervised contrastive learning~\citep{khosla2020supervised},
applying augmentations based on medical records, \eg~in the form of multiple views~\citep{vu2021medaug, azizi2021big},
learning from images and their reports~\citep{zhang2020contrastive},
applying SSL methods to adapt pretrained ImageNet models to the CXR domain~\citep{sowrirajan2021moco, gazda2021self, reed2022self},
and even a review of SSL applications in the medical domain~\citep{krishnan2022self}.
Our work is similar to the multi-stage training strategy of~\cite{reed2022self}, but we use \emph{unlabeled} \emph{generalist} pretraining on an out-of-domain dataset and \emph{unlabeled} specialist pretraining on an in-domain dataset.
\citet{azizi2021big} also perform SSL~PT on ImageNet followed by SSL~PT on CheXpert.
Interestingly, while they obtain small improvements (with 5 labels), we see that supervised learning surpasses SSL PT when using all labels, while the two come close when using 5 labels.

\begin{table}[t]
\tabcolsep=0.08cm
\small
\centering
\caption{Details of Chest X-Ray datasets.}
\vspace{-2mm}
\label{tab:datasets}
\begin{tabular}{lcc}
\toprule
Dataset & \# Samples & \# Labels \\
\midrule
NIH-CXR~{\tiny\citep{summers2019nih}} & 112,120 & 11 \\
% PadChest~{\tiny\citep{bustos2020padchest}} & 160,868 & 13 \\
CheXpert~{\tiny\citep{irvin2019chexpert}} & 224,316 & 12 \\ \bottomrule
\end{tabular}
\vspace{-3mm}
\end{table}

\section{Experimental Setup}
\label{sec:setup}

We present the datasets, methods, and implementation details used in this study.

\mypara{Datasets.}
% \label{subsec:datasets}
We formulate Chest X-Ray pathology detection as a multi-label classification problem, where the list of labels is specific to each dataset.
We perform experiments primarily on NIH-CXR~\citep{wang2017chestx, summers2019nih} and
% PadChest~ \citep{bustos2020padchest} and 
CheXpert~\citep{irvin2019chexpert},
where the list of labels in each dataset (post combination based on inputs from a radiologist) is presented in \tableref{tab:labels} (Appendix~\ref{apd:first}).
We split the data into 80:10:10 between train, validation and test while ensuring subjects are disjoint.

\mypara{SSL Methods.}
\label{subsec:ssl}
We demonstrate results on 5 different SSL algorithms.
We apply a limited set of augmentations to CXR images: horizontal flipping and rotation, as the data is already in grayscale, and addition of noise and blur may negatively affect performance~\citep{sowrirajan2021moco}.
We will see that even with these few and simple augmentations, pretraining on in-domain data shows comparable performance.

(i)~\textbf{SimCLR}~\citep{chen2020simple} is among the first SSL contrastive learning approaches. It aims to maximize similarity between representations of two augmentations of the same image in the latent space via a contrastive loss function, while all other images in the minibatch are treated as negative pairs.
% for the loss function.
(ii)~\textbf{MoCo}~\citep{he2020momentum} also uses two augmented views of an image but pairs them with two encoders, where
the parameter updates for the momentum encoder are performed through a linear interpolation between the two encoders.

The previous methods apply contrastive learning at an instance level and use negative samples in their formulation.
Different from them,
(iii)~\textbf{SwAV}~\citep{caron2020unsupervised} creates multiple clusters to partition the dataset, and attempts to map all augmented views of the same image to the same cluster, called prototype.
(iv)~\textbf{BYOL}~\citep{grill2020bootstrap} uses two networks - online and target - which have the same architecture. Here, the target network is used to teach the online network to correctly predict an augmented view of the same image.
Finally, (v)~\textbf{SimSiam}~\citep{chen2021exploring} uses two parallel encoders to generate representations of augmented views of the same image. However, gradient propagation is prevented in one encoder through the use of the Stop-Gradient operator.

\mypara{Fine-tuning data subsets.}
\label{subsec:train_details}
Our PT models are fine-tuned on different proportions of the NIH or CheXpert training sets.
We use subsets 1\%, 10\%, or 100\% of the training set to analyze the impact of fine-tuning on smaller subsets of the data.
For SSL PT, we use the entire 100\% NIH training set.

\begin{table*}[t]
\small
\centering
\caption{Comparison between models pretrained with different paradigms on NIH and CheXpert (CheX).
During PT, NIH corresponds to 100\% of the train set.
Supervised PT uses a standard classification setup for ImageNet (ImNet).
The number in \textcolor{forestgreen}{green} is the best result with supervised ImageNet PT, and the number in \textcolor{blue}{blue} is the best SSL PT strategy.}
\vspace{-2mm}
\label{tab:nih_unsupervised}
\begin{tabular}{@{}cccccccccc@{}}
\toprule
 & \multirow{3}{*}{Algorithm} & \multicolumn{2}{c}{Pre-training Dataset} & \multicolumn{6}{c}{Finetuning Dataset} \\
 &  & \multirow{2}{*}{Supervised} & \multirow{2}{*}{Unsupervised} & NIH & NIH & NIH & CheX & CheX & CheX \\
 &  &  &  & 1\% & 10\% & 100\% & 1\% & 10\% & 100\% \\ \midrule
1 & - & - & - & 0.5566 & 0.6794 & 0.7886 & 0.5582 & 0.6480 & 0.7534 \\
2 &  & ImNet & - & 0.6851 & 0.8026 & 0.8538 & 0.6916 & 0.7589 & 0.8066 \\ \midrule
3 & \multirow{4}{*}{SimCLR} & - & ImNet & 0.5818 & 0.6917 & 0.8297 & 0.5738 & 0.7123 & 0.7902 \\
4 &  & - & NIH & 0.5916 & 0.7481 & 0.8291 & \textcolor{blue}{0.6497} & 0.7314 & 0.7833 \\
5 &  & - & ImNet → NIH & 0.6109 & 0.7585 & 0.8435 & 0.6117 & \textcolor{blue}{0.7471} & 0.7990 \\
6 &  & ImNet & NIH & \textcolor{forestgreen}{0.7151} & \textcolor{forestgreen}{0.8097} & 0.8559 & \textcolor{forestgreen}{0.7170} & \textcolor{forestgreen}{0.7665} & \textcolor{forestgreen}{0.8089} \\ \midrule
7 & \multirow{4}{*}{SwAV} & - & ImNet & 0.5711 & 0.7031 & 0.8293 & 0.5643 & 0.7158 & 0.7933 \\
8 &  & - & NIH & 0.5880 & 0.7207 & 0.8384 & 0.6004 & 0.7328 & 0.7911 \\
9 &  & - & ImNet → NIH & \textcolor{blue}{0.6304} & 0.7591 & \textcolor{blue}{0.8496} & 0.6142 & \textcolor{blue}{0.7468} & \textcolor{blue}{0.8000} \\
10 &  & ImNet & NIH & 0.6347 & 0.8035 & 0.8540 & 0.6523 & 0.7569 & 0.8072 \\ \midrule
11 & \multirow{4}{*}{BYOL} & - & ImNet & 0.5861 & 0.7219 & 0.8362 & 0.5744 & 0.7289 & 0.7938 \\
12 &  & - & NIH & 0.6017 & 0.7154 & 0.8305 & 0.5786 & 0.7272 & 0.7889 \\
13 &  & - & ImNet → NIH & 0.5918 & \textcolor{blue}{0.7651} & 0.8445 & 0.6045 & 0.7449 & 0.7996 \\
14 &  & ImNet & NIH & 0.6783 & 0.7779 & 0.8496 & 0.6798 & 0.7614 & 0.8052 \\ \midrule
15 & \multirow{4}{*}{MoCo v2} & - & ImNet & 0.5711 & 0.7087 & 0.8359 & 0.6033 & 0.7324 & 0.7977 \\
16 &  & - & NIH & 0.5841 & 0.719 & 0.8041 & 0.6373 & 0.7085 & 0.7607 \\
17 &  & - & ImNet → NIH & 0.6258 & 0.7343 & 0.8379 & 0.6382 & 0.7417 & 0.7992 \\
18 &  & ImNet & NIH & 0.7061 & 0.8093 & \textcolor{forestgreen}{0.8569} & 0.7142 & \textcolor{forestgreen}{0.7666} & 0.8077 \\ \midrule
19 & \multirow{4}{*}{SimSiam} & - & ImNet & 0.5192 & 0.5630 & 0.7434 & 0.5503 & 0.6507 & 0.7659 \\
20 &  & - & NIH & 0.5143 & 0.5833 & 0.7852 & 0.5241 & 0.639 & 0.7586 \\
21 &  & - & ImNet → NIH & 0.5358 & 0.7039 & 0.8028 & 0.6244 & 0.7091 & 0.7668 \\
22 &  & ImNet & NIH & 0.5835 & 0.7796 & 0.8478 & 0.6714 & 0.7542 & 0.8036 \\ \bottomrule
\end{tabular}
\end{table*}

\mypara{Training details.}
Images are resized to $224 \times 224$ resolution for training.
We use a ResNet50 backbone followed by a linear layer for all our multi-label classification models, and adopt the Binary Cross-Entropy loss.
We perform ImageNet pretraining using~\citep{mmselfsup2021} and follow the same settings for NIH.
% After unsupervised pretraining, first on the ImageNet dataset (1000 classes) and then on NIH C,
Following this, we perform supervised fine-tuning on different CXR  datasets - NIH and CheXpert (CheX).
The models are fine-tuned for 30 epochs, with a learning rate of 1e-4, decayed by half every 5 epochs, and the Adam optimizer.
Hyperparameters are tuned on the validation set.

\mypara{Zero-shot experiments.}
We evaluate models fine-tuned on one dataset (\eg~NIH) on another dataset (\eg~CheXpert) to understand the zero-shot transfer capability.
As necessary, we restrict to the set of commonly used 5 labels~\citep{irvin2019chexpert}.

\mypara{Metrics.}
We report the mean AUROC to compare results for all experiments as in past literature (\eg~\citep{rajpurkar2018deep}).

\section{Results and Discussion}
\label{sec:discussion}

Table~\ref{tab:nih_unsupervised} reports results for various combinations of pretraining and fine-tuning strategies on both NIH and CheXpert datasets.

\begin{figure*}[t]
\centering
\includegraphics[width=0.75\linewidth]{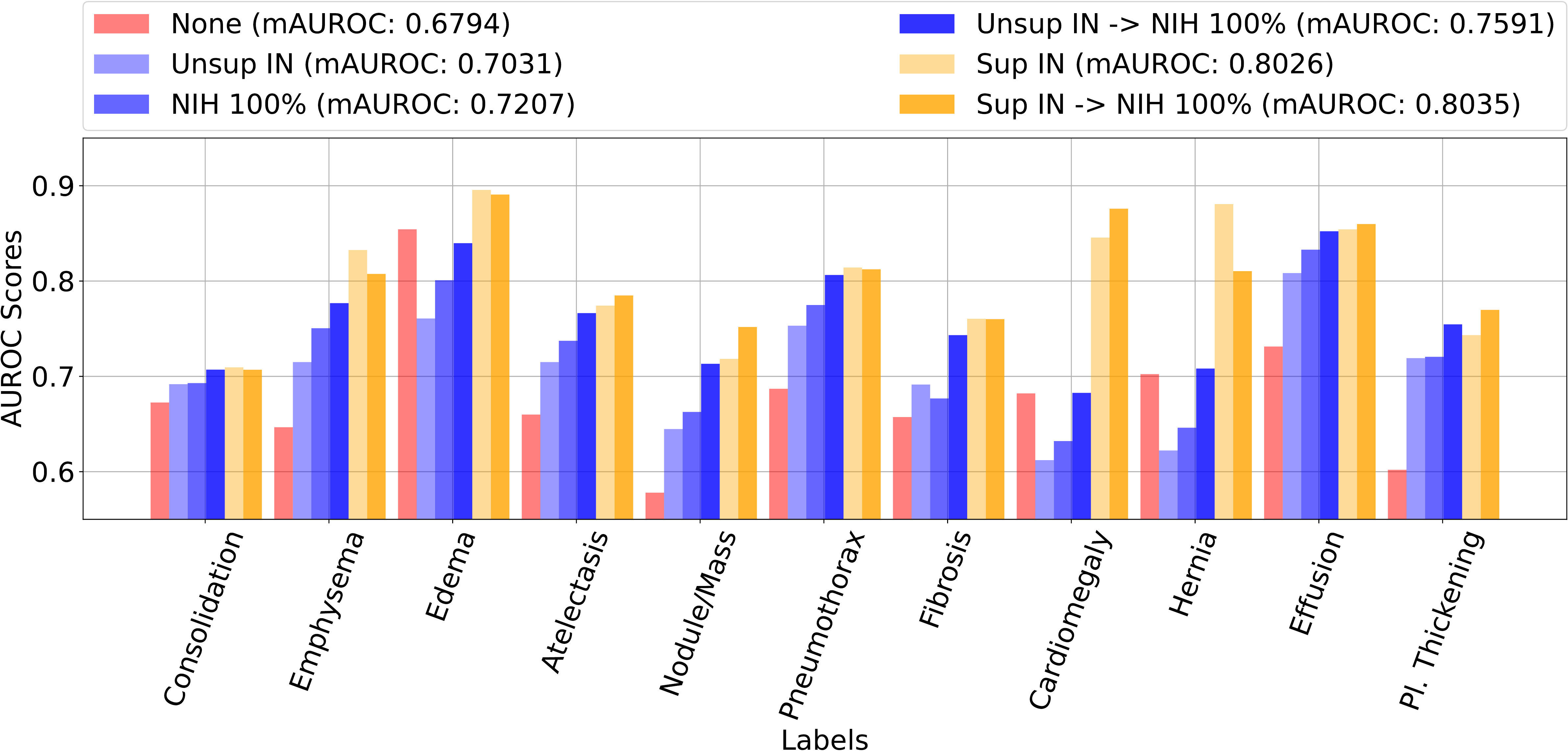}
% }
\vspace{-2mm}
\caption{Per-label AUROC. SSL PT algorithm: SwAV, Finetuning: NIH~10\% dataset.}
\label{fig:label_analysis}
\vspace{-4mm}
\end{figure*}

\mypara{Supervised PT}
on ImageNet, an out-of-domain dataset, strongly benefits model performance.
The difference in AUROC when training from scratch (row 1) \vs using ImageNet initialization (row 2) is 13-14\% (absolute AUROC points) for the small data subsets of NIH~1\% and CheX~1\%.
While this reduces progressively, there is a notable 5-7\% gap even for the entire training set (NIH~100\% or CheX~100\%).

\mypara{Does SSL ImageNet PT help?}
Yes.
Rows 3, 7, 11, and 15 show a consistent 3-5\% performance improvement over row 1 (training from scratch).
Interestingly SimSiam (row 19) is the only SSL method that hurts performance -- perhaps the domain gap is too large for the stop-gradient based training with parameter updates.

\mypara{Comparing NIH \vs ImageNet PT.}
Pairs of rows 3-4, 7-8, 11-12, and 15-16 allow us to compare the impact of in-domain (NIH) \vs out-of-domain (ImageNet) pretraining.
Note that ImageNet has about 1M images while NIH has less than 100K images.
We see largely comparable performance or small improvements of 0.5-1\% when fine-tuning on NIH, indicating the efficiency of a smaller in-domain PT dataset.
Importantly, in-domain PT shows a substantial improvement of 3-8\% over a model trained from scratch (row 1).
For CheXpert, while SimCLR, SwAV, and BYOL show 3-7\% improvements for CheX~1\%, when using the full CheX~100\% for training, ImageNet PT models are comparable or better.

\mypara{Does chaining SSL~PT strategies improve performance?}
We compare ImNet → NIH (rows 5, 9, 13, 17) against their individual ImageNet only or NIH only variants.
Barring a few exceptions, we see consistent improvements ranging from 0.5-5\% by chaining the PT strategies for all methods.
This verifies that the hierarchical pretraining strategy suggested by~\cite{reed2022self} is also applicable when both PT datasets are used in the self-supervised mode.
With this method, and when using 100\% of the fine-tuning datasets, we observe that SSL~PT models are less than 1\% away from supervised ImageNet PT (row 2).

\mypara{Is it possible to chain SSL in-domain PT with supervised ImageNet PT?}
Yes, in fact, row 6 (SimCLR) seems to achieve the best performance across 5 of the 6 settings, with row 18 (MoCo v2) being a close second in 2 settings.
We see larger improvements of 3\% and 2.6\% on the NIH~1\% and CheX~1\% subsets, however, these shrink when using the entire training set to 0.2-0.3\%.

\mypara{Which SSL method is the best?}
As expected, there is not one method that performs best in all scenarios.
However, SwAV seems to perform well in the SSL only settings, while SimCLR chains well with supervised ImageNet achieving good performance across all settings.
SimSiam underperforms on all fine-tuning results.

\mypara{Cross-dataset evaluation.}
As seen in the comparison between in-domain and out-of-domain PT, we note that the PT dataset is always NIH even when we fine-tune on CheXpert.
It is encouraging to see that PT on a different CXR image dataset still helps improve performance on CheXpert.

\mypara{Labelwise performance.}
Fig.~\ref{fig:label_analysis} shows the AUROC for individual labels when fine-tuned on the NIH~10\% dataset.
It is encouraging to observe that chaining unsupervised ImageNet and NIH PT (dark blue bar) outperforms SSL~PT on individual datasets across each label.
However, the same cannot be said for the supervised ImageNet settings (yellow bars).
Looking at the individual labels, we see a large variation in performance - interestingly, this is not only driven by the number of instances that are available for each pathology.

\begin{figure}[t]
\centering
\includegraphics[trim={0 4mm 0 0},clip,width=\linewidth]{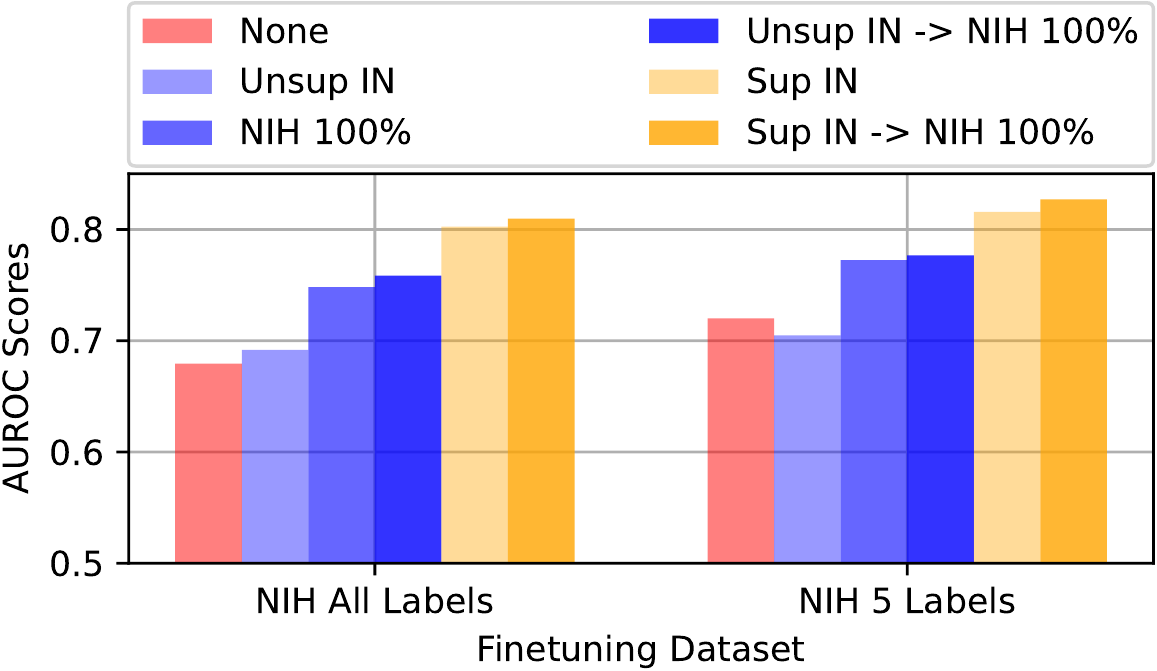} \vspace{2mm}
\includegraphics[trim={0 4mm 0 17.1mm},clip,width=\linewidth]{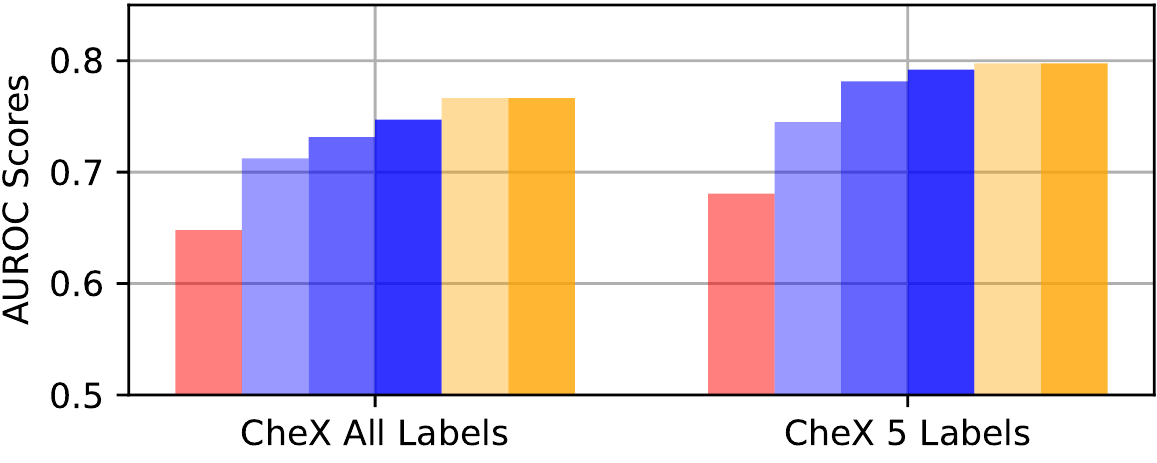}
\vspace{-9mm}
\caption{Comparing fine-tuning on 5 \vs \emph{all} labels using the SimCLR algorithm. Top: NIH, Bottom: CheXpert.}
\label{fig:5vsfull}
\vspace{-4mm}
\end{figure}

\mypara{Comparison of 5 \vs all labels.}
Fig.~\ref{fig:5vsfull} shows the results when evaluating on 5 or all labels.
As expected, all results improve when looking at a subset of 5 labels that appear often or are more important~\citep{irvin2019chexpert}.
Interestingly, the gap between the models trained using the SSL~PT strategy (blue bars) and supervised ImageNet models (light yellow bar) reduces as we transition from All to 5 labels for CheXpert.
We are not sure why this may happen.

\begin{figure}[t]
\centering
\includegraphics[trim={0 0 0 0},clip,width=\linewidth]{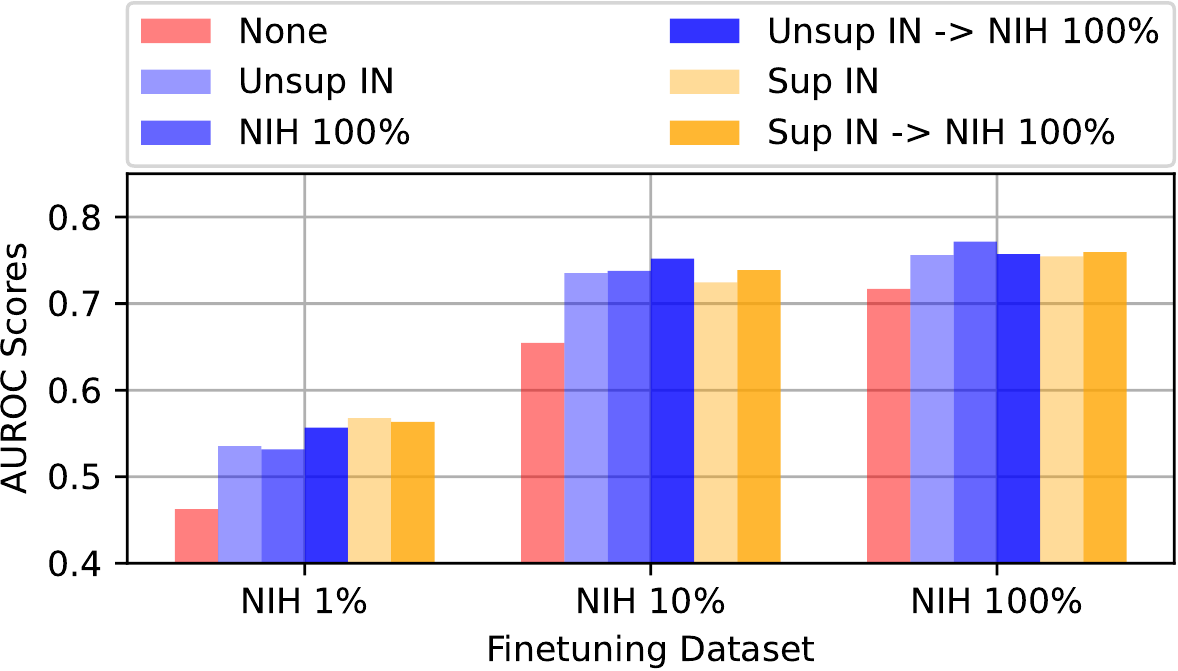} % \vspace{2mm}
\includegraphics[trim={0 0 0 17.1mm},clip,width=\linewidth]{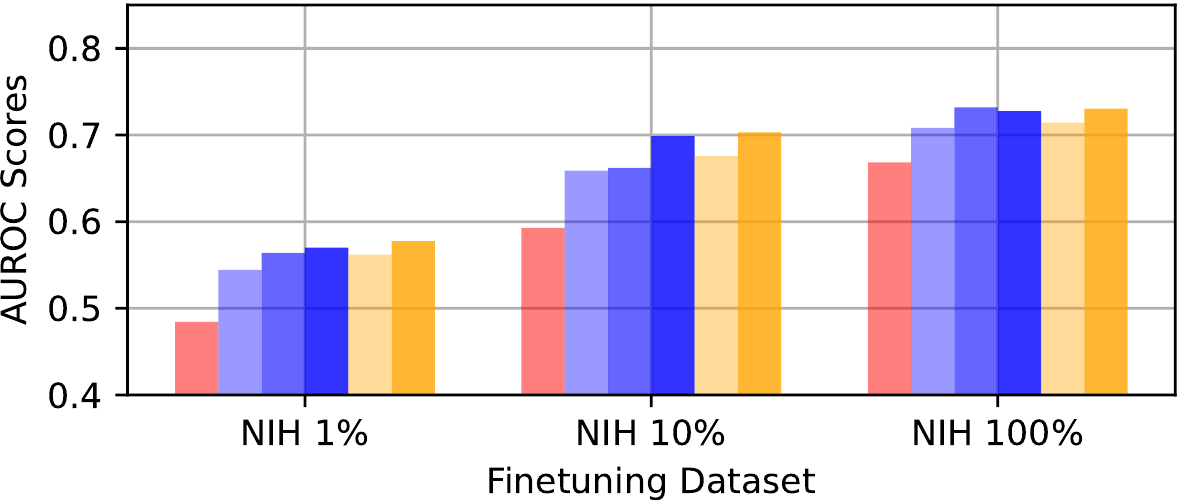} % \vspace{2mm}
\includegraphics[trim={0 0 0 17.1mm},clip,width=\linewidth]{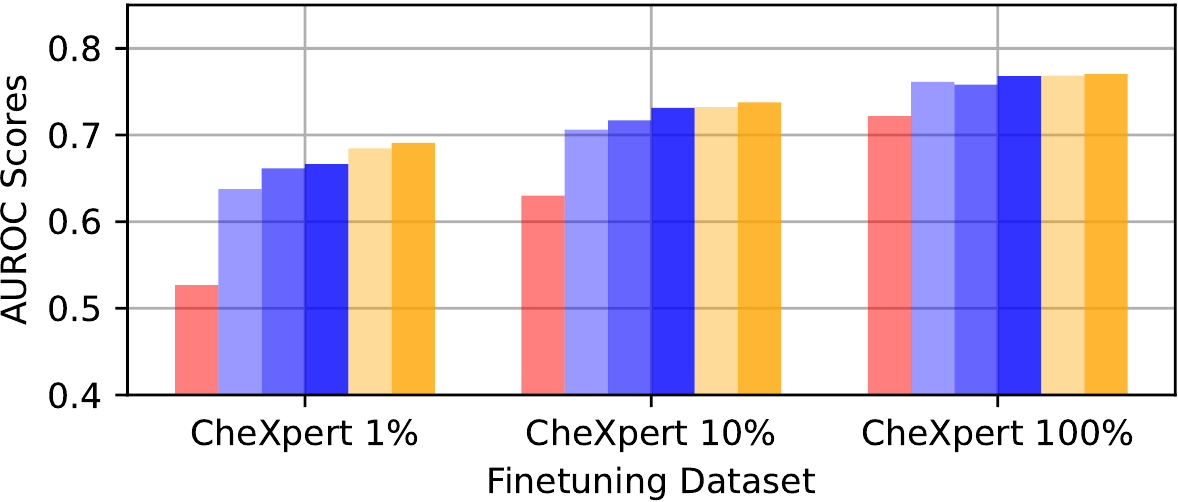}
\vspace{-6mm}
\caption{Zero-shot evaluation.
Models are PT using the SwAV algorithm.
Top: NIH → MIMIC (5 labels).
Middle: NIH → CheXpert (5 labels).
Bottom: CheXpert → MIMIC (all labels).}
\label{fig:zero_shot}
\vspace{-4mm}
\end{figure}

\mypara{Zero-shot evaluation.}
Finally, we present an experiment where fine-tuned models are evaluated across datasets.
In Fig.~\ref{fig:zero_shot}, we show results for models fine-tuned on NIH data on MIMIC-CXR~\citep{johnson2019mimic} and CheXpert (using 5 labels as there are different labels in these datasets), and a model fine-tuned on CheXpert evaluated on MIMIC with all labels (as they have the same label set).
We observe that models that have been trained with in-domain SSL~PT methods (dark blue and blue bars) outperform even supervised ImageNet PT.
This indicates that such models may be more robust to deploy in real-world scenarios as they are less affected by domain shifts in the dataset.

\mypara{Conclusion.}
We evaluated various supervised and self-supervised pretraining strategies for CXR datasets and showed the effects of in-domain and out-of-domain pretraining.

{\small
\mypara{Acknowledgements.}
This work is made possible by the generous support of the American people through the United States Agency for International Development (USAID). The contents are the responsibility of Wadhwani AI and do not necessarily reflect the views of USAID or the United States Government.
}

% \acks{Acknowledgements go here.}

\bibliography{refs}

\clearpage
\appendix 
\section{Dataset Details}\label{apd:first}

\begin{table}[t]
\tabcolsep=0.12cm
\small
\caption{Final labels in each dataset. Underlined labels are used in the experiments with 5 labels.}
\vspace{-4mm}
\label{tab:labels}
\begin{tabular}{@{}cc@{}}
\toprule
\multicolumn{1}{c}{\textbf{Dataset}} & \multicolumn{1}{c}{\textbf{Labels}} \\ \midrule
\begin{tabular}[c]{@{}l@{}}NIH-CXR\\ (11 labels)\end{tabular} & \begin{tabular}[c]{@{}l@{}}
\underline{Consolidation}, Emphysema,  \\ 
\underline{Edema}, \underline{Atelectasis}, Fibrosis, \\ 
Pneumothorax, Nodule/Mass, \\ 
\underline{Cardiomegaly}, \underline{Effusion},\\ 
Hernia, Pleural Thickening \end{tabular} \\ 
\midrule
\begin{tabular}[c]{@{}l@{}}CheXpert\\ (12 labels)\end{tabular} & \begin{tabular}[c]{@{}l@{}}
\underline{Consolidation}, \underline{Effusion},  \\ 
\underline{Edema}, Pneumothorax, \\ 
Nodule/Mass, Fracture, \\ 
Enlarged Cardiomediastinum,\\ 
Lung Opacity, Pleural Other,\\ 
\underline{Cardiomegaly}, \underline{Atelectasis},  \\
Support Devices \end{tabular} \\ 
% \midrule
% \begin{tabular}[c]{@{}l@{}}PadChest\\ (13 labels)\end{tabular} & \begin{tabular}[c]{@{}l@{}}Consolidation, Emphysema, \\ 
% Edema, Atelectasis, \\ 
% Nodule/Mass, Pneumothorax,\\ 
% Fibrosis, Cardiomegaly, \\ 
% Hernia, Effusion, Fracture,\\ 
% Pleural\_Thickening, \\ 
%  Air\_Trapping\end{tabular} \\ 
\bottomrule
\end{tabular}
\end{table}

Pathologically similar dataset labels were merged after consulting radiologists. This makes the dataset and our model more suitable for clinical deployment. 

NIH-CXR originally contained 14 labels. We’ve merged two sets of labels - 1) Infiltration, Consolidation and Pneumonia into Consolidation, and 2) Nodule and Mass into Nodule/Mass. For CheXpert, we 1) combined Consolidation and Pneumonia, 2) renamed Lung Lesion to Nodule/Mass to maintain consistency across datasets. The no finding label was removed, as absence of all labels in the multi-label setup automatically indicates the same. The final list of labels in each dataset is shown in (\cf~Table~\ref{tab:labels})

\section{Supplementary Results}\label{apd:second}

\begin{figure}[t]
\centering
\includegraphics[trim={0 4mm 0 0},clip,width=\linewidth]{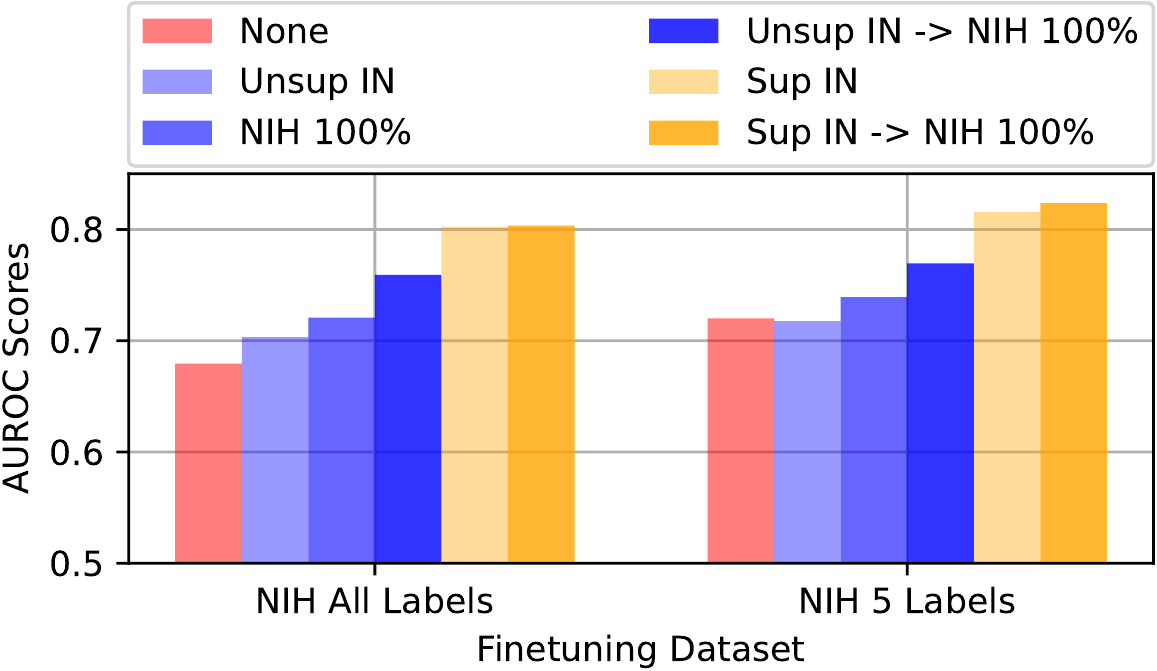} % \vspace{2mm}
\includegraphics[trim={0 4mm 0 17.1mm},clip,width=\linewidth]{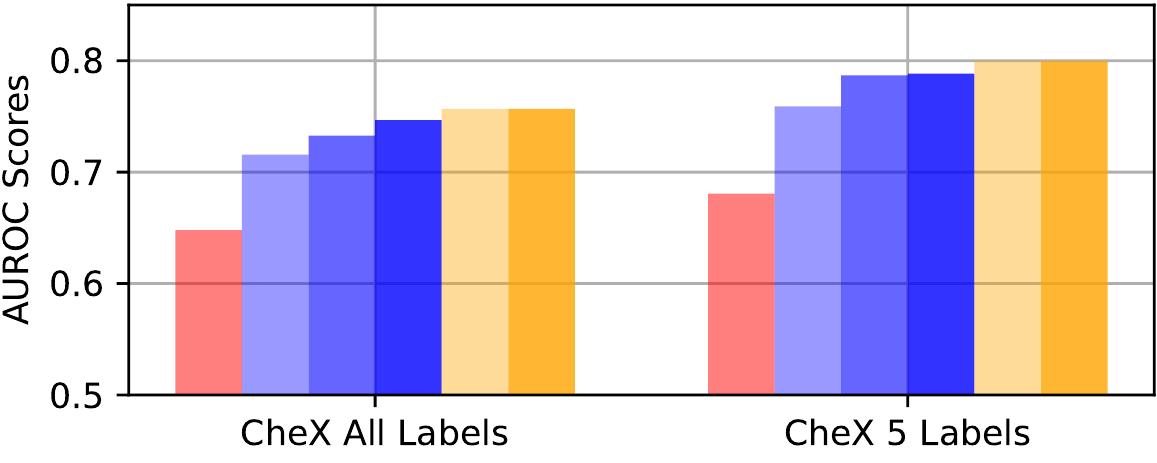}
\vspace{-5mm}
\caption{Comparing fine-tuning on 5 \vs \emph{all} labels using the SwAV algorithm. Top: NIH, Bottom: CheXpert.}
\label{fig:5vsfull_supplementary}
\vspace{-5mm}
\end{figure}

\mypara{Comparison of 5 \vs all labels.}
We see a trend similar to SimCLR from Fig.~\ref{fig:5vsfull} for the SwAV algorithm in Fig.~\ref{fig:5vsfull_supplementary}. While the SSL PT models finetuned on NIH 5 labels still lag behind the supervised models as in the all labels setting, the SSL PT models finetuned on 5 labels of CheXpert have a similar performance to those supervised PT models. This provides further evidence for the trend observed earlier.

\begin{figure}[t]
\centering
\includegraphics[trim={0 0 0 17.1mm},clip,width=\linewidth]{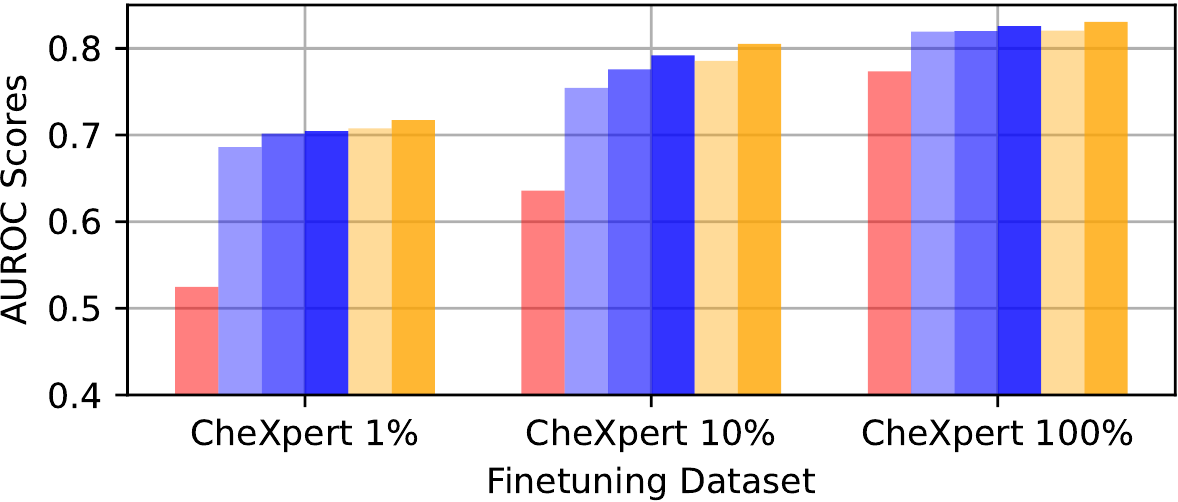}
\vspace{-5mm}
\caption{Zero-shot evaluation.
Models are PT using the SwAV algorithm.
CheXpert → NIH (5 labels).}
\label{fig:zero_shot_supp}
\vspace{-5mm}
\end{figure}

\mypara{Zero-shot evaluation.}
In Fig.~\ref{fig:zero_shot_supp}, we present the results of models finetuned on CheXpert and evaluated on NIH. The SSL PT models (blue bars) are particularly strong in this setting. Chained pre-training on unlabeled ImageNet and NIH either performs at par or slightly improves the results of the supervised models. This shows the promise of pre-training on unlabeled domain data, and then fine-tuning on labeled domain data, even if it has a different labeled set.

\end{document}